\documentclass[fleqn,10pt]{wlscirep}
\usepackage[utf8]{inputenc}
\usepackage[T1]{fontenc}
\usepackage{multirow}
\usepackage{lineno}
\title{Spatial frequency information fusion network for few-shot learning}

\author[1,*]{Wenqing Zhao}
\author[1,]{Guojia Xie}
\author[2]{Han Pan}
\author[2,3,*]{Biao Yang}
\author[1,*]{Weichuan Zhang}
\affil[1]{School of Electronic Information and Artificial Intelligence, Shaanxi University of Science and Technology, Xi’an 710000, China; 1668207599@qq.com, x311801@126.com, zwc2003@163.com}
\affil[2]{Society of Entrepreneurs and Ecology (SEE) Foundation, Beijing 100020, China (H.P.); panhan\underline{ }1996@163.com}
\affil[3]{Key Laboratory of Southwest China Wildlife Resources Conservation (Ministry of Education), China West Normal University, Nanchong 637002, China; yangb315@163.com}

\begin{abstract}
The objective of Few-shot learning is to fully leverage the limited data resources for exploring the latent correlations within the data by applying algorithms and training a model with outstanding performance that can adequately meet the demands of practical applications. In practical applications, the number of images in each category is usually less than that in traditional deep learning, which can lead to over-fitting and poor generalization performance. Currently, many Few-shot classification models pay more attention to spatial domain information while neglecting frequency domain information, which contains more feature information. Ignoring frequency domain information will prevent the model from fully exploiting feature information, which would effect the classification performance. Based on conventional data augmentation, this paper proposes an SFIFNet with innovative data preprocessing. The key of this method is enhancing the accuracy of image feature representation by integrating frequency domain information with spatial domain information. The experimental results demonstrate the effectiveness of this method in enhancing classification performance.
\end{abstract}
\begin{document}

\flushbottom
\maketitle
\thispagestyle{empty}

\noindent Please note: Few-shot learning; Frequency domain information; Spatial domain information; Data preprocessing; Data augmentation; Classification performance;

\section*{Introduction}

With the advancement of technology, data-driven deep learning \cite{ren4962361adaptive, r13, r399} models have achieved remarkable results in tasks across various fields. Intelligent algorithms such as deep learning models are employed to automate the monitoring of wildlife species by analyzing the massive amount of data collected \cite{r6}. However, these successes often hinge on large-scale labeled data. However, in many real-world scenarios, obtaining a large amount of labeled data is not easy. Annotated images for rare disease diagnosis are extremely hard to collect \cite{r3}. Humans are capable of learning stable feature representations with small training samples for dealing with image classification tasks. Inspired by this ability of
humans, few-shot learning has attracted much attention in computer vision and pattern recognition which intends to rapidly learn a classifier with good generalization capacity for understanding new concepts from very limited numbers of labeled training examples \cite{r36,tang2025cascading,du2025ccl,shui2013corner, li2024automotive,lei2024semi,lu2023track,jing2022recent,zhang2019corner, zhang2014corner, shui2012noise,zhang2019discrete, zhang2020corner,jing2022image,li2023traffic,zhang2023image,wang2020corner,lu2022image,jing2023ecfrnet}. Therefore, within the diverse subfield of deep learning, the value of few-shot learning \cite{wang2025principal, r19} is gradually on the rise. Concurrently, it is gradually gaining more attention from a larger number of people. However, at the same time, few-shot learning also faces numerous serious challenges. The existing few shot learning algorithms cannot fully extract the features of fine-grained images, leading to the low classification accuracy of fine-grained images \cite{r43}. For example, how can images be accurately classified when there are only a small number of labeled samples? Few-shot learning aims to solve the problems of the high cost of data annotation and difficult data acquisition in practical applications. It achieves high-precision classification by effectively utilizing limited data, to improve the ability of models to adapt to new tasks and its generalization performance. Especially in medical diagnosis, the ability to train accurate diagnostic models using limited labeled data is particularly applicable in the initial stage of collecting diagnostic data for emerging diseases. Few-shot learning techniques can reduce data scarcity issues and enhance medical image analysis \cite{r14} speed and robustness \cite{r1}. Collecting a substantial amount of training data poses a challenge for deep learning models in cancer diagnosis \cite{r4}. For example, when it comes to some rare types of cancer, few-shot learning models can help doctors identify and diagnose conditions more quickly and accurately. In the research area of newly discovered species, few-shot learning models contribute to the identification and protection of the species of the newest animals. In the field of medical imaging, obtaining a large amount of labeled data requires professional knowledge and a great deal of time, which incurs extremely high costs. Similarly, the distribution of animals in nature is inherently unbalanced, which causes some species to appear far more often than others \cite{r5,tang2024cascading}. In the field of rare species conservation, the number of these animals is extremely scarce, making it difficult to collect a large number of samples. In response to these issues, few-shot learning can train effective models even when there is only a small amount of labeled data.

Within the research scope of few-shot learning, metric learning has gradually become a hot topic. Metric-learning based methods intend to represent samples in an appropriate feature space where the feature information obtained from different categories can be distinguished based on similarity metrics \cite{r34}. The classic models with metric learning are shown below: DeepEMD \cite{r7} splits an image into multiple patches and then uses Earth Mover's Distance to calculate the optimal matching flow between the patches of the query set and the support set images. DeepBDC \cite{r8} calculates the inner product of the Brownian Distance Covariance matrices of a pair of images, and the similarity between the two images can be obtained. BSNet \cite{r9} consists of a single embedding module and a dual-similarity module. In the embedding module, the support images and query images first undergo operations based on convolutional neural networks and are mapped to a high-dimensional feature space and generate feature maps. Subsequently, these feature maps are fed into the dual-similarity module to calculate the Euclidean distance and the cosine distance to determine their categories.

The main difficulties faced by few-shot learning during the training process are severe model over-fitting and poor generalization ability \cite{r12}. This is mainly because the amount of data in many practical application scenarios is scarce. Few-shot learning techniques can reduce data scarcity issues and enhance medical image analysis, especially with meta-learning \cite{r2}. The essence of deep learning is using a sufficient amount of image data to optimize model parameters, imposing adequate constraints to make necessary adjustments. Weichuan Zhang and his team are the first to discover and analyze that existing metric-learning-based FSFGIC algorithms are prone to bias in the feature information obtained from each input image due to image quantization noise \cite{r38}. When the number of samples is too small, the model will over-fit the noise and local features in the training data, treating this non-universal information as patterns to learn. As a result, when faced with new data, the model is unable to generalize accurately, which leads to severe over-fitting. For example, in the scenario of a few-shot disease diagnosis using medical images. Due to the limited case data of rare diseases, the model may overly focus on certain special textures or markings in the training images, and may even include the noise generated during the image acquisition process in its learning scope. When encountering new cases, the features extracted previously do not commonly exist in the images of other patients because there are some differences between the new samples and the training samples. Since the model can't accurately identify the new features it leads to an increase in the misdiagnosis rate of new cases. This indicates that few-shot learning usually struggles to cover all the variations within a category, making it difficult for the model to learn comprehensive feature patterns. Therefore, when a few-shot learning model encounters a testing set that is slightly different from the training set \cite{r22}, there will be a significant decline in its performance.

In the process of exploring solutions to the dilemmas of few-shot learning, frequency domain information \cite{r15} has presented us with new opportunities for breakthroughs. When applied to few-shot learning, frequency domain transformation can convert data from the spatial domain into the frequency domain. So, it can unearth the unique features of the data in the frequency domain which may not be readily apparent in the spatial domain. Since frequency domain information is insensitive to local changes in data, it can represent the essential characteristics of data more stably of few-shot samples. The integration of frequency domain information offers a new research direction for solving the challenges of few-shot learning. In response to the difficulties of few-shot learning, this paper proposes a novel feature augment method \cite{r30} that involves integrating frequency-domain information into the few-shot learning methods that previously focused only on the spatial domain. This method aims to enhance the ability of few-shot learning models to extract features from limited sample data and improve their classification performance. When processing the input images, apply the Discrete Cosine Transform (DCT) \cite{r16, r28} method to obtain the frequency domain images. Then, extracting the low-frequency information of the frequency domain images, which contains the main structural and texture features of the images. Then, reconstructing the low-frequency information through the Inverse Discrete Cosine Transform (IDCT) \cite{r17} and adding it to the original images to obtain the enhanced images. This process effectively highlights the key features of the images. Meanwhile, it suppresses the interference of high-frequency noise, enabling the model to capture the essential features of the images more accurately. After experimental verification on multiple image classification datasets, our method has improved the classification accuracy and demonstrated superior performance compared with existing techniques.

\section*{Materials and Methods}

In few-shot learning, there is a small-scale support set and a query set that awaits classification. We generally refer to it as an n-way k-shot task, which indicates that the supporting dataset contains n different image categories, and each category consists of k-labeled samples. In few-shot learning, the number of samples in the support set is limited. However, it provides the feature information from various categories for the model, enabling the model to learn the feature patterns, differences, and commonalities of images in different categories. During the training process, the model will refer to the features and labels of the support set samples for learning and adjustment optimization. In the learning process, the model gradually develops a clear cognition of each category, which lays a solid foundation for classifying the samples in the query set later. The query set is composed of unlabeled image samples. It shares the same data feature and label space as the support set. After the model training is completed, the model needs to classify the images in the query set. After training, the model classifies samples in the query set by comparing and matching their features with those of various categories learned from the support set. Based on certain classification strategies, such as calculating distance metrics \cite{r18} or similarities, the model determines the category to which each query set sample belongs.

The network model proposed in this paper is evaluated on three datasets, CUB-200-2011 \cite{r20}, Stanford Dogs \cite{r21}, and a custom dataset containing 30 classes of animals. It should be noted that all the images in these datasets are collected via optical sensors. The CUB-200-2011 dataset includes 200 species of birds, with a total of 11,788 samples; the Stanford Dogs dataset covers 120 breeds of dogs and has 20,580 samples; the custom dataset of 30 classes of animals contains images of 10 classes of birds and 20 classes of mammals, with approximately 2000 samples in total. To ensure the effectiveness and rationality of model training, for the CUB-200-2011 dataset, we divide it into a training set (100 classes), a validation set (50 classes), and a test set (50 classes). The Stanford Dogs dataset is divided into a training set (70 classes), a validation set (20 classes), and a test set (30 classes). For the custom dataset of 30 classes of animals, 16 classes are selected for the training set, and 7 classes are included in each of the validation set and the test set. The training set needed to cover a broad range of categories to allow the model to learn diverse features and build a comprehensive knowledge base \cite{r41}. During the training process, the validation set is used to evaluate the performance of the model after each epoch. It has a relatively small number of classes, but the selected classes are representative. The results of the validation set are used to adjust the hyperparameters of the model. The test set is used to ultimately evaluate the overall performance of the model, and the selection of its classes is based on the requirement of comprehensiveness. The test set is generally neither overly complex nor overly simple to ensure the model's applicability in real-world scenarios. The table \ref{tab:table1} is shown below:

\begin{table}[htbp]
    \centering
    \caption{Table of Model Performance Comparison Based on CUB-200-2011, Stanford Dogs, and a Custom Dataset of 30 Animal Classes}
    \label{tab:table1}
    \begin{tabular}{cccc}
        \hline
        \textbf{Dataset} & \textbf{\(N_{\text{train}}\) } & \textbf{\(N_{\text{val}}\) } & \textbf{\(N_{\text{test}}\) } \\
        \hline
        CUB-200-2011 & 100 & 50 & 50 \\
        Stanford Dogs & 70 & 20 & 30 \\
        Animals & 16 & 7 & 7 \\
        \hline
    \end{tabular}
\end{table}

On the above three datasets, we conducted experiments under the 5-way 1-shot 
\cite{r23} and 5-way 5-shot \cite{r23} few-shot learning. In this study, all experiments were run on a computer equipped with an Nvidia A6000 GPU using Ubuntu 22.04 and CUDA 11.8. The SFIFNet uses the PyTorch 1.7.1 framework, the other four models use the PyTorch 2.2.0 framework. We used ResNet12 and ViT \cite{r24} as the backbone to obtain feature representations. The Stochastic Gradient Descent (SGD) \cite{r25} algorithm was selected as the optimizer, with an initial learning rate of 0.001. To prevent the model from over-fitting, we set the weight decay \cite{r26} coefficient to 0.0005. By adding a regularization term \cite{r27} to the loss function, we constrained the weight parameters of the model, enabling the model to focus more on generalization ability when learning data features. In terms of the training strategy, we adopted a method of adjusting the learning rate in stages. The entire training process consists of 400 epochs. After a certain number of epochs, the learning rate is multiplied by 0.1 for decay. At each stage, the learning rate decreases by an order of magnitude, gradually reducing from 0.001. This learning rate decay strategy enables the model to converge rapidly in the early stage of training and fine-tune the parameters more precisely in the later stage. In the validation phase, we conduct validation on the standard 5-way 1-shot and 5-way 5-shot tasks every 20 epochs to monitor the model's performance on the validation set in real-time. We randomly generated 1000 tasks and obtained the average classification accuracy result by calculating the average accuracy of these 1000 tasks.

First, perform the DCT on the dataset. DCT decomposes an images into a combination of cosine waves of different frequencies, transforming the images from the spatial domain to the frequency domain. This transformation enables the model to capture the internal laws of the data from the perspective of frequency, revealing patterns and features that are not easily detectable in the spatial domain. DCT involves multiplying each pixel value f(x,y) of the images by cosine functions in two directions, and then summing up the results, which is essentially calculating the amplitudes of the cosine components of the images at different frequencies.

After the DCT transformation, we can concentrate most of the images's energy on the low-frequency coefficients. The main structural and contour information of images is usually contained in the low-frequency part, while the high-frequency part contains a large amount of detailed information or noise information. The low-frequency coefficients are usually located in the upper left corner of the matrix after DCT transformation. These coefficients represent the smooth changes and main shapes of the images. High-frequency coefficients are distributed in the lower right corner of the matrix. They correspond to rapid changes and details in the images. In few-shot learning, due to the limited number of samples, the model is prone to be disturbed by changes in image details or noise. Since low-frequency features are relatively stable, we can set a threshold to set the lower right part of the matrix to 0, which retains the low-frequency part and discard the high-frequency part. High-frequency filtering enables the model to better capture the key features of the images, reducing the impact of insufficient samples and changes in details.

Subsequently, perform the IDCT on the images with only low-frequency information retained, and then conduct information fusion \cite{r29}. IDCT combines cosine waves of different frequencies in the frequency domain to reconstruct the images in the spatial domain. Then, directly add the reconstructed images to the original images to enhance the main features of the images. The images that have undergone the high-frequency filtering operation still contain the main structural and contour information but may have lost some details. After converting this low-frequency information back to the spatial domain through the IDCT, adding it to the original images can not only highlight the key structures of the images but also replenish the detailed information in the original images.

In gray-level images, edges may be defined as sharp changes in intensity and most common types include steps, lines, and junctions \cite{r40}. So besides directly obtaining low-frequency information by cropping the upper left corner of the frequency domain images, we also considered a method of partitioning low-frequency components with gradients \cite{r31}. In the frequency domain, the low-frequency region generally corresponds to the smooth parts of the images, where the gradient values are small and the changes are slow. In contrast, the high-frequency region corresponds to the edges and detailed parts of the images, where the gradient values are large and the changes are drastic. First, we calculate the gradients of the DCT coefficient matrix in the horizontal and vertical directions and compute the gradient amplitudes. These gradient values reflect the degree of spatial variation of pixel values. Then, we calculate the mean and standard deviation of the gradient amplitudes and then determine a threshold based on the mean and standard deviation. By setting a gradient threshold, the regions where the gradient values are smaller than the threshold can be considered as low-frequency regions which is going to be retained. The coefficients with gradient amplitudes greater than the threshold are set to 0. However, it has been abandoned because of the slow running speed.

\section*{Results}

\subsection*{Structure of Spatial Frequency Information fusion Processing Flow}

Explanation of the flow: The original images are duplicated into two copies. One copy is used to preserve the original spatial information, and the other undergoes DCT to obtain its frequency-domain information. Subsequently, a low-pass filter is applied to obtain the frequency-domain information of images. With IDCT, low-frequency images are obtained which is going to be fused with the images that preserve the original spatial information. Then the fused images are fed into the backbone composed of ResNet12 and ViT for subsequent processing. The network structure is shown in the following figure \ref{fig:net_structure}:

\begin{figure}[htbp]
    \centering
    \includegraphics[width=1\textwidth]{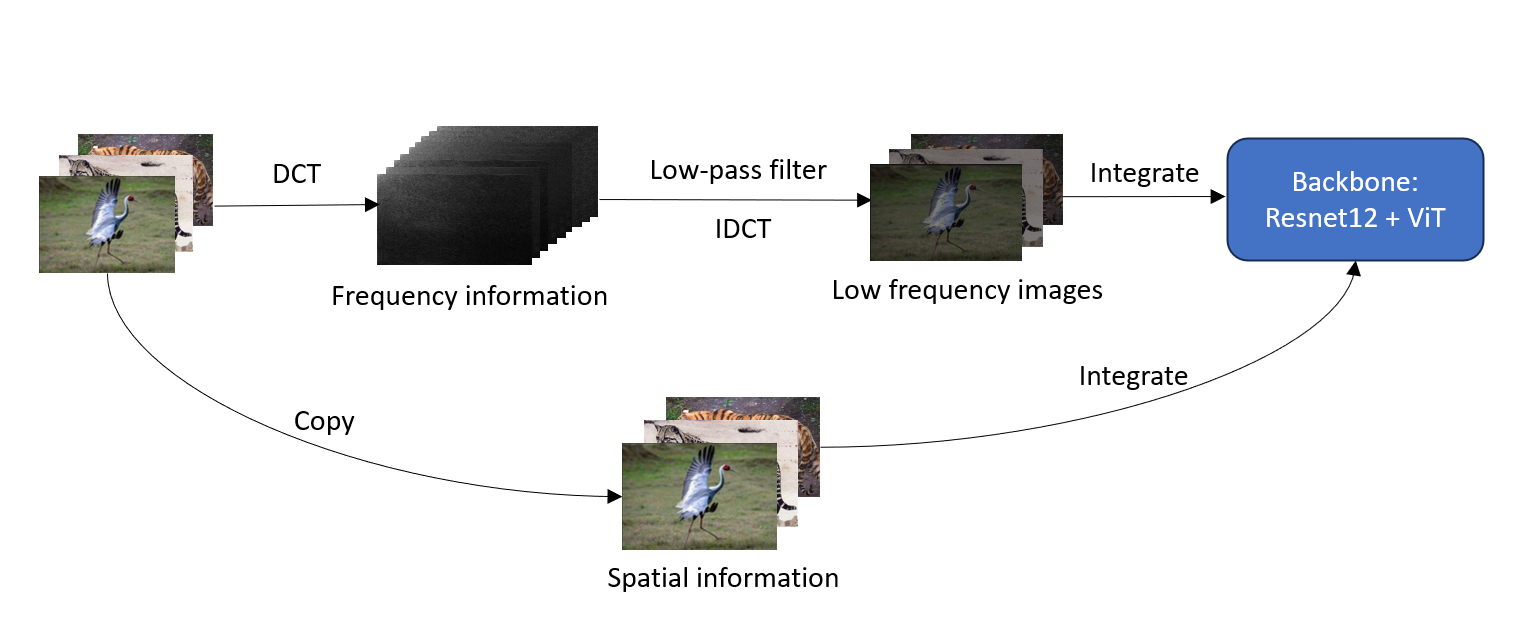}
    \caption{Flow of extracting low-frequency information in the frequency domain and performing information fusion}
    \label{fig:net_structure}
\end{figure}

\subsection*{Performance comparison}

Datasets have become one of the most critical roles in the development of FSFGIC, not only as a means for evaluating the classification accuracy of different FSFGIC methods, but also for greatly promoting the development of the field of FSFGIC (e.g., solving more complex, practical, and challenging problems) \cite{r37}. In this section, we comprehensively compared and evaluated the performance of the SFIFNet for extracting frequency domain information against classic few-shot images classification networks such as FRN \cite{r32}, BSNet, DeepEMD, and DeepBDC on the CUB-200-2011, Stanford Dogs, and a custom dataset containing 30 classes of animals. The backbone of the SFIFNet is ResNet12 and ViT. The backbone selected for FRN, DeepEMD, and DeepBDC is ResNet12. The backbone of BSNet is CNN \cite{r33}. All experiments were carried out under the 5-way 1-shot and 5-way 5-shot settings respectively. Judging from the experimental results, the SFIFNet demonstrated outstanding performance on each dataset, significantly outperforming the FRN, BSNet, DeepEMD, and DeepBDC networks. The results are shown in the following table \ref{tab:table2}:

\begin{table}[htbp]
    \centering
    \caption{Table of Model Performance Comparison Based on CUB-200-2011, Stanford Dogs, and a Custom Dataset of 30 Animal Classes}
    \label{tab:table2}
    \resizebox{\linewidth}{!}{
    \begin{tabular}{cccccccc}
        \hline
        \multirow{2}{*}{Methods} & \multicolumn{2}{c}{CUB-200-2011} & \multicolumn{2}{c}{Stanford Dogs} & \multicolumn{2}{c}{Animals} \\
        \cline{2 - 7}
        & 1 - Shot & 5 - Shot & 1 - Shot & 5 - Shot & 1 - Shot & 5 - Shot \\
        \hline
        BSNet & 62.84 ± 0.95  & 85.39 ± 0.56 & 43.42 ± 0.86 & 71.90 ± 0.68 & 32.36 ± 0.62 & 40.50 ± 0.54 \\
        DeepBDC & 81.98 ± 0.44 & 92.24 ± 0.24 & 73.57 ± 0.46 & 86.61 ± 0.27 & 34.31 ± 0.33 & 48.66 ± 0.31 \\
        DeepEMD & 75.59 ± 0.30 & 88.23 ± 0.18 & 70.38 ± 0.30 & 85.24 ± 0.18 & 33.97 ± 0.21 & 45.47 ± 0.47 \\
        FRN & 83.94 ± 0.19 & 93.77 ± 0.10 & 77.89 ± 0.21 & 89.10 ± 0.12 & 59.12 ± 0.18 & 77.73 ± 0.10 \\
        ours & 93.66 ± 0.13 & 97.78 ± 0.05 & 83.90 ± 0.20 & 92.45 ± 0.12 & 75.92 ± 0.17 & 91.65 ± 0.07 \\
        \hline
    \end{tabular}
    }
\end{table}

From the experimental results of the CUB-200-2011 dataset and the Stanford Dogs dataset, we can conclude that the SFIFNet has a significant accuracy advantage in both 5-way 1-shot and 5-way 5-shot tasks. For the custom dataset of animals, the accuracy of the SFIFNet in the 5-way 1-shot and 5-way 5-shot experiments far exceeds that of FRN, BSNet, DeepEMD, and DeepBDC. This fully verifies that the SFIFNet not only performs excellently on public datasets but also has a strong generalization ability for custom datasets with a small amount of data.

\subsection*{Ablation Studies}

To accurately evaluate the effectiveness of adding a preprocessing procedure for extracting frequency domain information to the ResNet12 + ViT backbone, we conducted ablation experiments on the CUB-200-2011, Stanford Dogs, and a custom animals dataset under the 5-way 1-shot and 5-way 5-shot few-shot learning. The experiment mainly focused on comparing the performance of models with and without the spatial frequency information fusion for extracting frequency domain information, so as to clarify the specific contribution of this processing procedure to the model's performance. The results are shown in the following table \ref{tab:table3}:

\begin{table}[htbp]
    \centering
    \caption{The influence of whether to add a preprocessing process for extracting frequency-domain information on the network performance.}
    \label{tab:table3}
    \resizebox{\linewidth}{!}{
        \begin{tabular}{ccccccc}
            \hline
            \multirow{2}{*}{Status} & \multicolumn{2}{c}{CUB-200-2011} & \multicolumn{2}{c}{Stanford Dogs} & \multicolumn{2}{c}{Animals} \\
            \cline{2-7}
            & 1-Shot & 5-Shot & 1-Shot & 5-Shot & 1-Shot & 5-Shot \\
            \hline
            Unmodified & 92.31 ± 0.21 & 96.67 ± 0.14 & 81.05 ± 0.32 & 89.75 ± 0.29 & 72.89 ± 0.26 & 88.37 ± 0.24 \\
            \hline
            SFIFNet & 93.66 ± 0.13 & 97.78 ± 0.05 & 83.90 ± 0.20 & 92.45 ± 0.12 & 75.92 ± 0.17 & 91.65 ± 0.07 \\
            \hline
        \end{tabular}
    }
\end{table}

Judging from the experimental results, in the 5-way 1-shot and 5-way 5-shot experiments conducted on all datasets, the accuracy of the SFIFNet is slightly higher than that of the original model.

\subsection*{Mathematical components explain the process of frequency information extraction}

In the preprocessing procedure for extracting frequency information, the DCT equation is as follows:

\begin{linenomath}
\begin{equation}
F(u, v) = \frac{1}{4} C(u) C(v) \sum_{x=0}^{N-1} \sum_{y=0}^{M-1} f(x, y) \cos \left[ \frac{(2x+1)u\pi}{2N} \right] \cos \left[ \frac{(2y+1)v\pi}{2M} \right].
\end{equation}
\end{linenomath}

\(F(u,v)\) is the coefficient matrix after DCT transformation. \(f(x, y)\) is the original images matrix. \(N\) and \(N\) are the width and height of the images respectively. C(u) and C(v) are normalization coefficients, defined as:

\begin{linenomath}
\begin{equation}
C(u) = \begin{cases} 
\frac{1}{\sqrt{2}} & \text{if } u = 0 \\
1 & \text{if } u \neq 0 
\end{cases}, \space
C(v) = \begin{cases} 
\frac{1}{\sqrt{2}} & \text{if } v = 0 \\
1 & \text{if } v \neq 0 
\end{cases}
\end{equation}
\end{linenomath}

In the DCT equation, \(u\) and \(v\) represent the indices of the horizontal and vertical frequency components in the frequency domain respectively, ranging from 0 to \(N-1\) and \(M-1\). When \(u = 0\) and \(v = 0\), it represents the direct current component, which is the average brightness of the images. As \(u\) and \(v\) increase, the represented frequency components become higher. By selecting different values of \(u\) and \(v\), we can obtain the DCT coefficients of the images at different frequencies.
Subsequently, we extract the low-frequency part of the frequency domain images. The equation is as follows:

\begin{linenomath}
\begin{equation}
L_{i,j} = \begin{cases} 
D_{i,j} & \text{if } i < \lfloor h \sqrt{r} \rfloor \text{ and } j < \lfloor w \sqrt{r} \rfloor \\
0 & \text{otherwise},
\end{cases}
\end{equation}
\end{linenomath}

\(D\) is the coefficient matrix after DCT transformation. L is the coefficient matrix after extracting the low-frequency information. The \(h\) and \(w\) represent the height and width of the DCT coefficient matrix respectively. The \(r\) is the retention ratio of the low-frequency information, which is 0.15. $\lfloor x\rfloor$ represents the floor function, which returns the largest integer not greater than $x$. This equation indicates that we only retain the top-left $\lfloor hr\rfloor\times\lfloor wr\rfloor$ coefficients of the DCT coefficient matrix, and set all the remaining coefficients to 0. In this way, we can extract the low-frequency information of the images, preparing for the IDCT transformation and image fusion.
Subsequently, we perform the IDCT on the images. The equation is as follows:

\begin{linenomath}
\begin{equation}
f(x, y) = \frac{1}{4} \sum_{u=0}^{N-1} \sum_{v=0}^{M-1} C(u) C(v) F(u, v) \cos \left[ \frac{(2x+1)u\pi}{2N} \right] \cos \left[ \frac{(2y+1)v\pi}{2M} \right].
\end{equation}
\end{linenomath}

\(f(x,y)\) is the grayscale value of the pixel at \((x,y)\) of the images after IDCT transformation in the spatial domain. \(N\) and \(M\) are the width and height of the images respectively. \(F(u,v)\) is the coefficient after DCT transformation, representing the amplitude of the frequency component at \((u,v)\) of the images in the frequency domain. \(C(u)\) and \(C(v)\) are normalization coefficients, which are defined as:

\begin{linenomath}
\begin{equation}
C(u) = \begin{cases} 
\frac{1}{\sqrt{2}} & \text{if } u = 0 \\
1 & \text{if } u \neq 0 
\end{cases}, \space
C(v) = \begin{cases} 
\frac{1}{\sqrt{2}} & \text{if } v = 0 \\
1 & \text{if } v \neq 0 
\end{cases}
\end{equation}
\end{linenomath}

$\cos \left[ \frac{(2x + 1)u\pi}{2N} \right]$ and $\cos \left[ \frac{(2y + 1)v\pi}{2M} \right]$ are the cosine components in the horizontal and vertical directions respectively, which are used to calculate the contribution of each pixel at different frequencies.
In this study, to extract the low-frequency information of the images more accurately, we tried a low-pass filtering method based on gradient information. The equation of the low pass filter can be expressed as follows:

\begin{linenomath}
\begin{equation}
L_{i,j} = \begin{cases} 
D_{i,j} & \text{if } \sqrt{\left(\frac{\partial D}{\partial x}\right)_{i,j}^2 + \left(\frac{\partial D}{\partial y}\right)_{i,j}^2} \leq \mu_G + 2\sigma_G \\
0 & \text{otherwise},
\end{cases}
\end{equation}
\end{linenomath}

\(D\) is the coefficient matrix after DCT transformation. \(L\) is the coefficient matrix after applying the low-pass filter. \(\left(\frac{\partial D}{\partial x}\right)_{i,j}\) and \(\left(\frac{\partial D}{\partial y}\right)_{i,j}\) are the gradients of the DCT coefficient matrix in the \(x\) and \(y\) directions at the position \((i,j)\) respectively. \(\mu_G\) is the mean value of the gradient amplitudes, and \(\sigma_G\) is the standard deviation of the gradient amplitudes. However, since this method requires calculating the gradient of each coefficient, the computational complexity is high, resulting in a slow running speed. This has become a significant bottleneck when dealing with large-scale datasets. Therefore, despite the advantages of this method in terms of effectiveness, considering the efficiency in practical applications, we abandoned this method and instead adopted a more efficient low-frequency extraction strategy. 
After the above preprocessing procedures, including DCT transformation, low-frequency information extraction, and fusion operation with the original images, the processed images are fed into the backbone to extract the deep features of the images.

\section*{Discussion}

In this paper, based on the network architectures of ResNet12 and ViT as the backbone networks, we constructed a preprocessing procedure for extracting frequency-domain information from input images. Specifically, when the original images are input into the network, we first duplicate them to obtain two identical images. One of the images is used to preserve the original spatial-domain detail features of the images. The other images enter the frequency-domain processing procedure, that is, they undergo a discrete cosine transform. Researchers found that the image frequency representation can be further characterized into low-frequency generalization information, which contains smooth contour of objects, and high-frequency discriminative information with finer edge details of the images \cite{r35}. By transforming the images from the spatial domain to the frequency domain and only retaining the low-frequency information while discarding the high-frequency part in the frequency domain, we can highlight the main structure and content of the images and remove interference factors such as noise. It is worth noting that during the process of retaining the low-frequency information, the size of the images needs to be maintained unchanged to ensure the feasibility of subsequent processing such as inputting it into the backbone network. Subsequently, the inverse discrete cosine transform (IDCT) is performed on the images with only low-frequency information retained, and it is directly added to the previously retained spatial-domain images, achieving an organic combination of frequency-domain and spatial-domain features. Then the fused images are sent into the backbone network. From the experimental results, we can conclude that after the fused images are fed into the backbone for deep feature extraction, more abundant and representative high-level semantic information is successfully mined. These operations not only effectively improve the ability of models to extract image features but also enhance the robustness and generalization ability of models in complex scenarios. 

To mitigate the problems caused by the insufficiency of few-shot learning data, conventional data augmentation techniques are widely applied. Data augmentation techniques play a significant positive role in alleviating the problem of data insufficiency. Operations such as rotation, cropping, scaling, and color jittering, for example, increase data diversity. They provide more samples with different perspectives and feature combinations for model training. In image classification tasks, the model can learn the feature representations of the same object at different positions and scales by performing data augmentation operations on training images, which improves the robustness of the model to changes in object position and size. However, this approach is not without flaws. It can cause quantization noise, such as quantization errors, color distortion, and geometric deformation, which could interfere with the original features of the images. Therefore, it learns both the true features of the images and artificially introduced noises when the backbone network extracts features. As a result, the features extracted by the model will deviate, and the classification performance will be affected.

Although traditional data augmentation increases data diversity, the interference of noise may affect the accuracy of feature representation. The quality of feature representation directly affects the classification performance on FSFGIC \cite{r39}. The organic integration of frequency-domain features and spatial-domain features provides more abundant information. As a result, the backbone network can extract more representative high-level semantic information. In addition, the model can learn image features from different dimensions, which enhances its robustness in complex scenarios. In acquisition and transmission, images are often corrupted by noise \cite{r42}. Even in the face of noise interference introduced by traditional data augmentation, the model can more accurately recognize images relying on the fused features, reducing the impact of noise on classification performance and better addressing the problems of insufficient data and data augmentation in few-shot learning. Therefore, in practical applications, such as in medical disease diagnosis, fusing frequency-domain information can more accurately extract key information like lesion features, which improves the accuracy of disease detection and the ability for early diagnosis. In the field of rare animal conservation, fusing frequency-domain information can assist in more clearly identifying and analyzing the image features of rare animals, making it more effective for monitoring, identification, and protection. 

\section*{Conclusion}
In this paper, we focus on the few-shot image classification task. Since the most common challenge in few-shot learning tasks is the lack of training data, the model is prone to over-fitting. It is difficult for the network to learn sufficient features to distinguish different categories, which affects the accuracy of classification. To address the mentioned issues, an innovative image preprocessing method is proposed. This method extracts the low-frequency information from the images after performing DCT on them and then fuses the frequency domain information with the spatial domain information. This method can effectively improve the quality of the images input into the backbone, enabling the backbone to extract discriminative features more efficiently and alleviating the adverse effects caused by the scarcity of data. The method we proposed has unique advantages. This preprocessing method has a relatively low computational cost and can be flexibly combined with the backbone. It can be easily integrated into various existing training networks. The experimental results show that, compared to other few-shot image classification methods, our method achieves better classification performance on multiple few-shot image classification datasets.

\bibliography{sample}

\section*{Acknowledgements (not compulsory)}

The dataset used in this research was kindly provided by H.P., which significantly supported the smooth progress of this study.

\section*{Author contributions statement}

Conceptualization, Weichuan Zhang; methodology, Guojia Xie, Weichuan Zhang and Wenqing Zhao; software, Wenqing Zhao; validation, Wenqing Zhao; formal analysis, B.Y.; investigation, H.P.; resources, H.P.; data curation, Wenqing Zhao and H.P.; writing---original draft preparation, Wenqing Zhao; writing---review and editing, Weichuan Zhang and Wenqing Zhao; visualization, B.Y.; supervision, Weichuan Zhang, H.P. and B.Y.; project administration, Weichuan Zhang, H.P., B.Y. and Wenqing Zhao; funding acquisition, H.P. All authors have read and agreed to the published version of the manuscript.

\section*{Data availability}
All data generated or analysed during this study are included in this published article.

\section*{Additional information}
\subsection*{Competing interests}
The authors declare no conflicts of interest.

\end{document}